\documentclass[letterpaper, 10 pt, journal, twoside]{ieeetran}

\IEEEoverridecommandlockouts                              % This command is only needed if 
\makeatletter
\let\NAT@parse\undefined
\let\mycaption\@makecaption
\makeatother

\usepackage[numbers,sort&compress]{natbib} % References
\usepackage{times}                         % Times font
\usepackage{graphicx}                      % Include graphics
\usepackage{amsmath,amssymb,mathtools,bm}  % Math packages
\usepackage[inline]{enumitem}              % Inline/compact lists
\usepackage{xspace}                        % Smart spacing after macros
\usepackage{balance}                       % Balance columns in the last page
\usepackage{pifont}                        % Special symbols (e.g., ding)
\usepackage{siunitx}

\usepackage{caption}
\usepackage{subcaption}                    % Modern replacement for subfigure
\usepackage[table]{xcolor}                 % Table coloring
\usepackage{adjustbox}                     % Resize tables/figures
\usepackage{multicol,multirow,booktabs}    % Table styling
\usepackage[normalem]{ulem}                % Underlines & strikeouts (without changing \emph)
\usepackage{array}
\newcolumntype{C}[1]{>{\centering\arraybackslash}p{#1}}
\usepackage{makecell}

\usepackage{algorithm}
\usepackage{algpseudocode}

\usepackage{soul}                          % Text highlighting
\usepackage{color}                         % Basic color support

\newcommand{\bl}[1]{\textcolor{blue}{#1}}
\definecolor{orangered}{HTML}{FF4500}
\definecolor{dodgerblue}{HTML}{1E90FF}
\definecolor{darkgreen}{HTML}{008000}

\usepackage[hidelinks]{hyperref}           % Clickable refs (load before cleveref)
\definecolor{lightblue}{rgb}{0.21,0.49,0.74}
\hypersetup{
    colorlinks=true,
    linkcolor=blue,
    citecolor=green,
    filecolor=magenta,
    urlcolor=magenta
}

\usepackage{cleveref}                      % Intelligent cross-references
\crefname{figure}{Fig.}{Figs.}
\crefname{table}{Table}{Tables}
\crefname{section}{Sec.}{Secs.}
\crefname{equation}{}{}

\usepackage{lipsum}                        % Dummy text (optional, for testing)

\usepackage{RPM_packages/RPM_notations}
\usepackage{RPM_packages/RPM_acronyms}

\title{
LAPS: Improving Incremental LiDAR Mapping using Active Pooling and Sampling for Neural Distance Fields
}

\newcommand{\nickname}{LAPS\xspace}
\author{Dongjae Lee$^{1}$, Wooseong Yang$^{1}$, Yifu Tao$^{2}$, Maurice Fallon$^{2}$, and Ayoung Kim$^{1*}$%
\thanks{Manuscript received: January, 22, 2026; Revised April, 8, 2026; Accepted May, 6, 2026.}%Use only for final RAL version
\thanks{This paper was recommended for publication by Editor Javier Civera upon evaluation of the Associate Editor and Reviewers’ comments.
This work was supported in part by the National Research Foundation of Korea (NRF) grant funded by the Korea government (MSIT) (No. RS-2023-00241758), Korea Institute for Advancement of Technology (KIAT) grant funded by the Korea Government (MOTIE) (P0020536, HRD Program for Industrial Innovation), and the EPSRC-funded project Mobile Robotic Inspector (EP/Z531212/1). Maurice Fallon is supported by a Royal Society University Research Fellowship. For the purpose of open access, the authors have applied a Creative Commons Attribution (CC BY) license to any Accepted Manuscript version arising.} %Use only for final RAL version
\thanks{$^{1}$D. Lee, W. Yang, and A. Kim are with the Department of Mechanical Engineering, Seoul National University, S. Korea.
        {\tt\footnotesize [pur22, yellowish, ayoungk]@snu.ac.kr}}%
\thanks{$^{2}$Y. Tao and M. Fallon are with the Oxford Robotics Institute at the University of Oxford, UK.
        {\tt\footnotesize [yifu, mfallon]@robots.ox.ac.uk}}%
\thanks{Digital Object Identifier (DOI): see top of this page.}
}

\begin{document}

\maketitle
% \thispagestyle{empty}
% \pagestyle{empty}

%%%% BODY TEXT
\begin{abstract}

Neural distance fields offer a compact and continuous representation of 3D geometry, making them attractive for incremental LiDAR mapping.
However, their online optimization is vulnerable to catastrophic forgetting, where new observations can degrade previously reconstructed geometry.
Replay-based training is commonly used to address this issue, but existing methods typically rely on passive replay buffers and uniform sampling, which can waste memory on redundant observations and under-train poorly constrained regions.
We propose \nickname, a replay management framework for incremental neural mapping that improves both replay retention and replay allocation during online updates.
\nickname combines reliability-based active pooling to retain reliable historical samples under limited memory with uncertainty-guided active sampling to focus optimization on under-constrained regions.
Experiments on synthetic and real-world benchmarks show that \nickname consistently improves reconstruction completeness while maintaining competitive geometric accuracy.
On Oxford Spires, it improves recall by 4.66 pp and F1-score by 3.79 pp over PIN-SLAM on the Blenheim Palace 05 sequence.
We release our open source implementation at: \href{https://github.com/dongjae0107/LAPS}{https://github.com/dongjae0107/LAPS}.

\end{abstract}

\section{INTRODUCTION}
\label{sec:intro}

\IEEEPARstart{D}{ense} 3D mapping is a fundamental capability in robotics, supporting downstream tasks such as planning, navigation, and localization~\citep{sh-ch5-map}.
For robust autonomy, maps must be both accurate and complete across large environments.
Conventional explicit representations, such as point clouds or voxel grids, can achieve high fidelity, but doing so at scale requires high spatial resolution, which is very memory intensive~\citep{zhong2023shine}.
This limitation has motivated growing interest in implicit neural representations~\citep{takikawa2021neural,muller2022instant}, particularly neural distance fields, which model scene geometry with a compact parameterization~\citep{pan2024pin}.

In many real-world deployments, however, mapping must be performed online in an incremental manner. 
In this setting, neural mapping is susceptible to catastrophic forgetting, where new observations can overwrite previously learned geometry and degrade reconstruction quality~\citep{zhong2023shine}.
A widely adopted solution is replay-based training, where historical supervision is stored in a \textit{replay buffer} and interleaved with current samples during online optimization~\citep{ortiz2022isdf,pan2024pin}.
Existing work on incremental neural LiDAR mapping has largely focused on scalable map architectures for large environments~\citep{zhong2023shine,pan2024pin}, while the replay mechanism, specifically how historical samples are retained (pooling) and revisited during online optimization (sampling), has received less attention despite its direct impact on mapping performance.

% Figure 1
%%%%%
\begin{figure}[!t]
    \centering
    \includegraphics[width=1.00\linewidth]{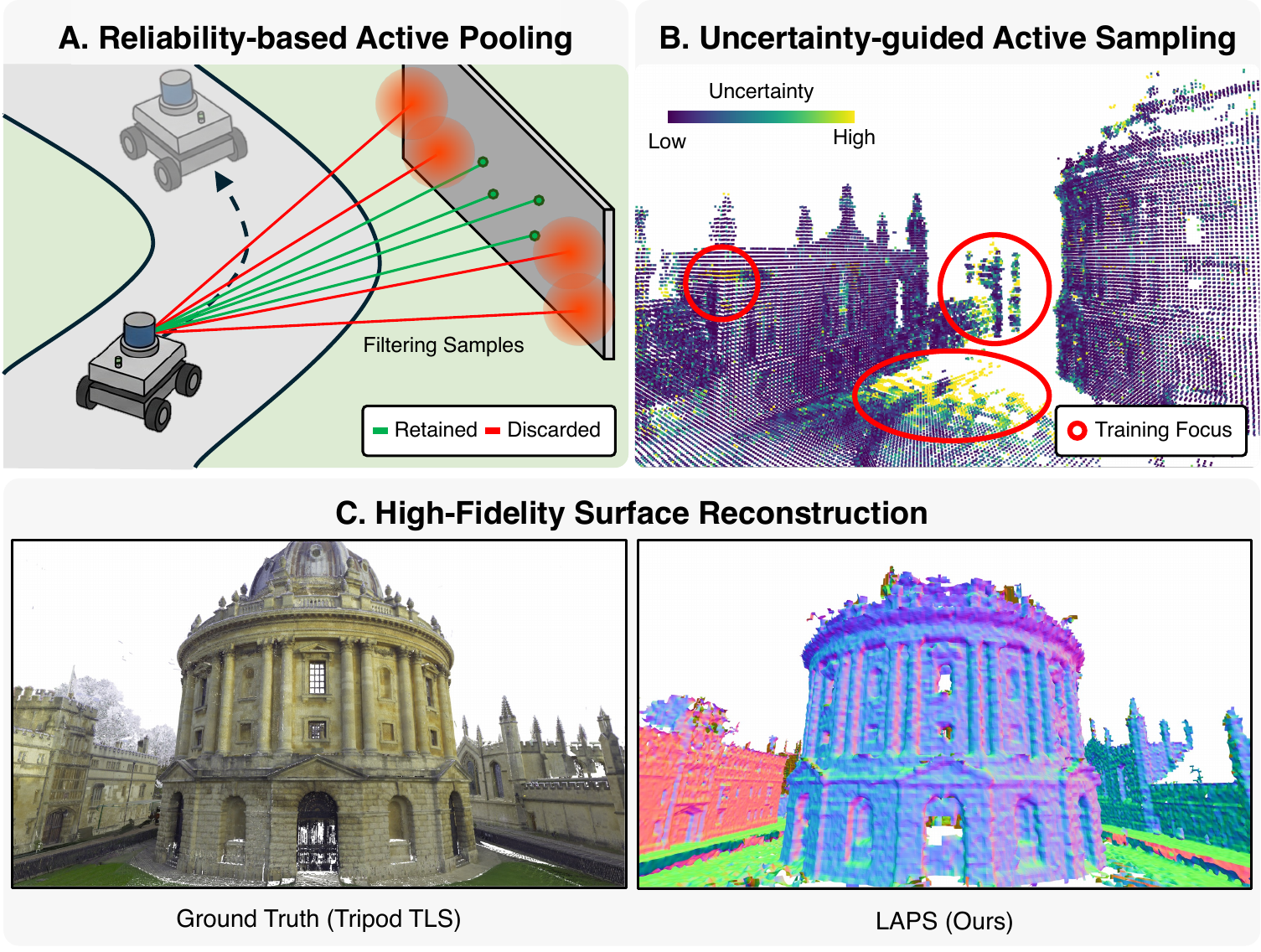}
    \caption{
    \textbf{Overview of \nickname.}
    \nickname incrementally reconstructs 3D environments from LiDAR scans with a neural distance field and manages online replay through (i) reliability-based active pooling to maintain a compact buffer with reduced spatial sample imbalance, and (ii) uncertainty-guided active sampling to focus optimization on under-constrained regions.
    Note that in the Radcliffe Camera example, the top region is not observed by the LiDAR and is captured only by the terrestrial laser scanner (TLS).
    }
    \label{fig:main}
    % \vspace{-1.0em}
\end{figure}

% \mfallon{modify figure 1 to say "Ground Truth (from Tripod Scanner)" - to make it clear that we cannot expect to reconstruct this. you might also want to mention that the top of the Radcliffe Camera was not actually scanned by the Hesai. Otherwise it asks the question why the top of the building is missing - when it is clear that there is line-of-sight to the top.}
%%%%%

We argue that replay management is a key yet underexplored bottleneck in incremental neural LiDAR mapping under online constraints.
With a fixed memory budget, the replay buffer must selectively retain past observations, but passive strategies, such as local windows~\citep{song2024n} or fixed-capacity queues~\citep{pan2024pin}, can over-represent frequently scanned regions while discarding supervision from sparsely observed areas.
With a limited online optimization budget, the method must also decide which retained samples receive more training effort.
Standard random sampling~\citep{pan2024pin} often revisits already well-constrained regions, leaving under-observed parts insufficiently optimized.
Consequently, reconstructions may remain locally accurate in dense regions but globally incomplete, with degraded surfaces or missing structures in under-constrained areas~\citep{song2024n}.

In this paper, we address incremental neural LiDAR mapping from the perspective of replay management under fixed memory and optimization budgets.
To this end, we propose \nickname, a unified replay management framework based on neural distance fields with two components (\cref{fig:main}).
First, reliability-based active pooling regulates replay storage by retaining the most reliable historical samples and reduces spatial sample imbalance in the replay buffer.
Second, uncertainty-guided active sampling uses estimated model uncertainty to prioritize under-constrained regions during online optimization, allocating the limited training budget more effectively.
Our contributions are:
\begin{itemize}
    \item We formulate replay management as a core challenge in incremental neural LiDAR mapping under fixed memory and online optimization constraints.
    \item We propose \nickname, a unified replay management framework that combines reliability-based active pooling for replay retention with uncertainty-guided active sampling for replay allocation.
    \item We demonstrate on synthetic and real-world benchmarks that \nickname improves reconstruction completeness while maintaining competitive geometric accuracy.
\end{itemize}
\section{RELATED WORK}
\label{sec:related}

\subsection{Map Representations}
\label{sub:inr}
Older dense mapping techniques in robotics have long relied on explicit representations such as point clouds and voxel grids.
While effective, they typically trade spatial resolution for memory efficiency, since preserving fine geometry at a large scale can become prohibitively expensive.
Prior work has improved scalability using sparse or adaptive volumetric structures, including voxel hashing~\citep{niessner2013real}, octrees~\citep{hornung2013octomap}, and VDBs~\citep{vizzo2022vdbfusion}.
Still, high-resolution explicit volumetric maps can require substantial memory in large environments.

Neural distance fields provide an alternative by representing geometry as a continuous \ac{SDF}, parameterized by a neural network~\citep{ortiz2022isdf}.
They support distance queries at arbitrary locations, enabling dense surface reconstruction with a compact representation.
To improve scalability, recent systems use hybrid parameterizations such as multi-resolution hash grids~\citep{muller2022instant} or hierarchical structures~\citep{zhong2023shine} together with neural decoders.
Building on these advances, several recent LiDAR-based mapping systems adopt neural distance fields as the map representation, with prior work focusing mainly on scalable neural architectures~\citep{zhong2023shine,pan2024pin}.
In incremental settings, however, performance also depends on replay management during online optimization, which has been less well explored.

\subsection{Incremental Neural Mapping}
\label{sub:incremental}
Incremental neural mapping is susceptible to catastrophic forgetting, where new supervision degrades previously learned geometry~\citep{zhong2023shine}.
Regularization-based approaches~\citep{zhong2023shine} aim to preserve past structure by constraining parameter updates, but their additional computational cost can hinder online operation.
As a result, many recent systems~\citep{sucar2021imap,ortiz2022isdf,pan2024pin,song2024n} rely on replay, training on both current and historical observations to mitigate forgetting.

The effectiveness of replay, however, is limited by buffer scalability.
Over long trajectories, replay-based methods must either store ever-growing data or discard past observations.
Early keyframe-based approaches~\citep{sucar2021imap,ortiz2022isdf} that retain all keyframes do not scale well to large environments, motivating more practical strategies such as local windows~\citep{pan2024pin,song2024n} or fixed-capacity buffers~\citep{pan2024pin}.
However, these passive policies can still be inefficient.
Window-based replay may accumulate redundant samples when the robot is moving slow or is stationary, while fixed-capacity queues may discard informative observations.
In addition, non-uniform point density can introduce spatial imbalance, causing densely observed regions to dominate both the replay buffer and subsequent training batches~\citep{song2024n}.
These limitations motivate replay management strategies that not only control memory usage but also prioritize which historical samples to retain and revisit during online learning.

\subsection{Uncertainty in Neural Reconstruction}
\label{sub:uncertainty}
Uncertainty modeling has been widely studied in neural implicit representations and 3D reconstruction~\citep{goli2024bayes,tao2025silvr,jiang2024fisherrf}.
A common distinction is between aleatoric uncertainty, which captures observation noise, and epistemic uncertainty, which reflects model uncertainty due to limited supervision.
In neural reconstruction, prior work has used aleatoric uncertainty for supervision weighting~\citep{song2025un3}, while epistemic uncertainty has mainly been used for filtering or confidence estimation~\citep{goli2024bayes,tao2025silvr}, or for active view selection in radiance field mapping~\citep{jiang2024fisherrf}.
In contrast, we use epistemic uncertainty to construct training batches in online incremental mapping, prioritizing under-constrained regions during optimization.
To the best of our knowledge, this formulation has not been explored previously.
\section{METHOD}
\label{sec:method}

% Figure - Pipeline
%%%%%
\begin{figure*}[!t]
    \centering
    \includegraphics[trim={0.3cm 0 0 0},clip, width=0.95\linewidth]{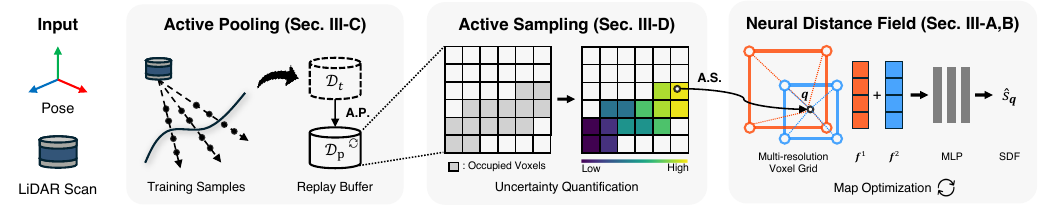}
    \caption{
    \textbf{\nickname pipeline.}
    At each time step, we convert the incoming LiDAR scan into TSDF supervision samples and use them to update a neural distance field.
    The samples are inserted into a replay buffer through reliability-based active pooling.
    We then repeatedly draw mini-batches using uncertainty-guided active sampling and perform optimization steps to update the neural map.
    }
    \label{fig:overview}
    % \vspace{-1.0em}
\end{figure*}
%%%%%

The pipeline of \nickname is shown in \cref{fig:overview}.
Given a stream of LiDAR scans with corresponding poses, the system incrementally optimizes a neural map that represents the scene as a continuous \ac{TSDF} (\cref{sub:map}).
At each time step, the incoming scan is converted into \ac{TSDF} supervision samples (\cref{sub:training}).
These samples are then incorporated into a replay buffer using reliability-based active pooling (\cref{sub:pooling}), and the model is updated through optimization iterations using mini-batches drawn by uncertainty-guided active sampling (\cref{sub:sampling}).

\subsection{Implicit Neural Map Representation}
\label{sub:map}
We represent the environment with an implicit neural distance field parameterized by a multi-resolution sparse voxel grid, following prior work~\citep{muller2022instant, zhong20243d}.
The grid contains $L$ resolution levels and allocates voxels on demand when they become occupied by LiDAR measurements.
At each level, a learnable feature vector \(\vf \in \Real{D}\) is stored at each of the eight corner vertices of a voxel, and spatial hash tables are maintained for efficient feature lookup~\citep{teschner2003optimized, muller2022instant}.

For an arbitrary 3D query point \(\vq \in \Real{3}\), we compute its feature \(\vf_\vq\) by aggregating trilinear interpolations of the surrounding voxel corner features across all resolution levels:
\begin{equation}
\label{eq:1}
    \vf_\vq = \sum_{l=1}^L \mathrm{Trilinear}(\vq, \{\vf^l_k\}_{k=1}^8),
\end{equation}
where \(\{\vf^l_k\}\) denotes the feature vectors at the eight corner vertices of the voxel enclosing \(\vq\) at level \(l\).
The aggregated feature \(\vf_\vq\) is then fed to a globally shared \ac{MLP} decoder \(\Phi\) to predict the truncated signed distance:
\begin{equation}
\label{eq:2}
    \hat{s}_\vq = \Phi(\vf_\vq).
\end{equation}
During incremental mapping, both the grid features and the decoder parameters are jointly optimized via backpropagation.

\subsection{Training Samples and Loss Function}
\label{sub:training}
At time step \(t\), the input scan is denoted as \(\calP_t = \{\vp_i\}_{i=1}^{N_t}\) with sensor origin \(\vo_t\). 
For each observed point \(\vp_i\), we define the ray from the origin to the point as \(\vr_i = \vp_i - \vo_t \). 
A sample point \(\vu\) at depth \(d\) along this ray is
\begin{equation}
\label{eq:3}
    \vu = \vo_t + \frac{d}{\| \vr_i \|_2}\vr_i.
\end{equation}
We generate \ac{TSDF} supervision by sampling points along each ray using three depth distributions. 
Specifically, \(N_f\) samples are drawn in front of the observed surface point with depths \(d_f \sim \mathcal{U}(\| \vr_i \|_2 - d_{tr}, \| \vr_i \|_2)\), and  \(N_b\) samples are drawn behind the surface with depths \(d_b \sim \mathcal{U}(\| \vr_i \|_2, \| \vr_i \|_2 + d_{tr})\), where \(d_{tr}\) denotes the truncation distance.
In addition, \(N_r\) free-space samples are drawn in front of the surface with depths \(d_r \sim \mathcal{U}(d_{min}, \| \vr_i \|_2 - d_{tr})\), where \(d_{min}\) is the minimum sampling depth.
Including \(\vp_i\), each scan yields \(N_{s} = N_t (1+N_f+N_b+N_r)\) training samples. 

Each sampled point is assigned a projective signed distance based on the difference between the surface depth and the sampled depth,
\begin{equation}
\label{eq:4}
    \tilde{s} = \| \vr_i \|_2 - d,
\end{equation}
which is then truncated to the \ac{TSDF} interval \([-d_{tr}, d_{tr}]\) and used as the supervision label. 
The resulting training set \(\mathcal{D}_t = \{(\vu, \tilde{s})\}\) is appended to a replay buffer \(\mathcal{D}_p\) to mitigate catastrophic forgetting (\cref{sub:pooling}).
For map optimization, a batch of \(N\) samples is sampled from \(\mathcal{D}_p\) (\cref{sub:sampling}).
The neural distance field is trained by regressing the truncated signed distance values using \bl{a} \ac{MSE} loss:
\begin{equation}
\label{eq:5}
    \mathcal{L} = \frac{1}{N} \sum_{i=1}^N (\tilde{s}_i - \hat{s}_i)^2,
\end{equation}
where \(\tilde{s}_i\) and \(\hat{s}_i\) denote the projective and predicted signed distances, respectively.

\subsection{Reliability-based Active Pooling}
\label{sub:pooling}
As described above, we maintain a training data pool \(\mathcal{D}_p\) to reuse historical supervision samples.
However, storing all samples is not memory-scalable and can induce severe spatial imbalance, where densely observed regions dominate the replay buffer and bias training~\citep{song2024n}.

To address this, we introduce a reliability-based active pooling strategy that keeps the buffer compact while reducing spatial sample imbalance. 
After appending \(\mathcal{D}_t\) to \(\mathcal{D}_p\), we first retain only samples \(\vu \in \mathcal{D}_p\) within a local window centered at the current sensor origin $\vo_t$,
\begin{equation}
\label{eq:6}
   \Vert \vu - \vo_t \Vert_2 < r_p,
\end{equation}
where \(r_p\) is the pooling radius.
Each remaining sample is then assigned to the voxel at the lowest grid resolution, yielding a set of voxels \(\{v_k\}\).

To reduce spatial sample imbalance, we cap each voxel \(v_k\) to at most \(\tau\) samples.
When \(\vert v_k \vert > \tau\), we retain the \(\tau\) most \textit{reliable} samples by ranking them using an estimated supervision error.
Concretely, we define the supervision error \(e\) as the difference between the projective signed distance \(\tilde{s}\) and the unknown true signed distance \(s\):
\begin{equation}
\label{eq:7}
   e = \tilde{s} - s \sim \mathcal{N}(\mathrm{b}, \sigma^2_a),
\end{equation}
where \(\mathrm{b}\) represents systematic bias in projective distance, and \(\sigma_a\) captures measurement uncertainty.

\begin{figure}[!t]
    \centering
    \includegraphics[width=0.99\linewidth]{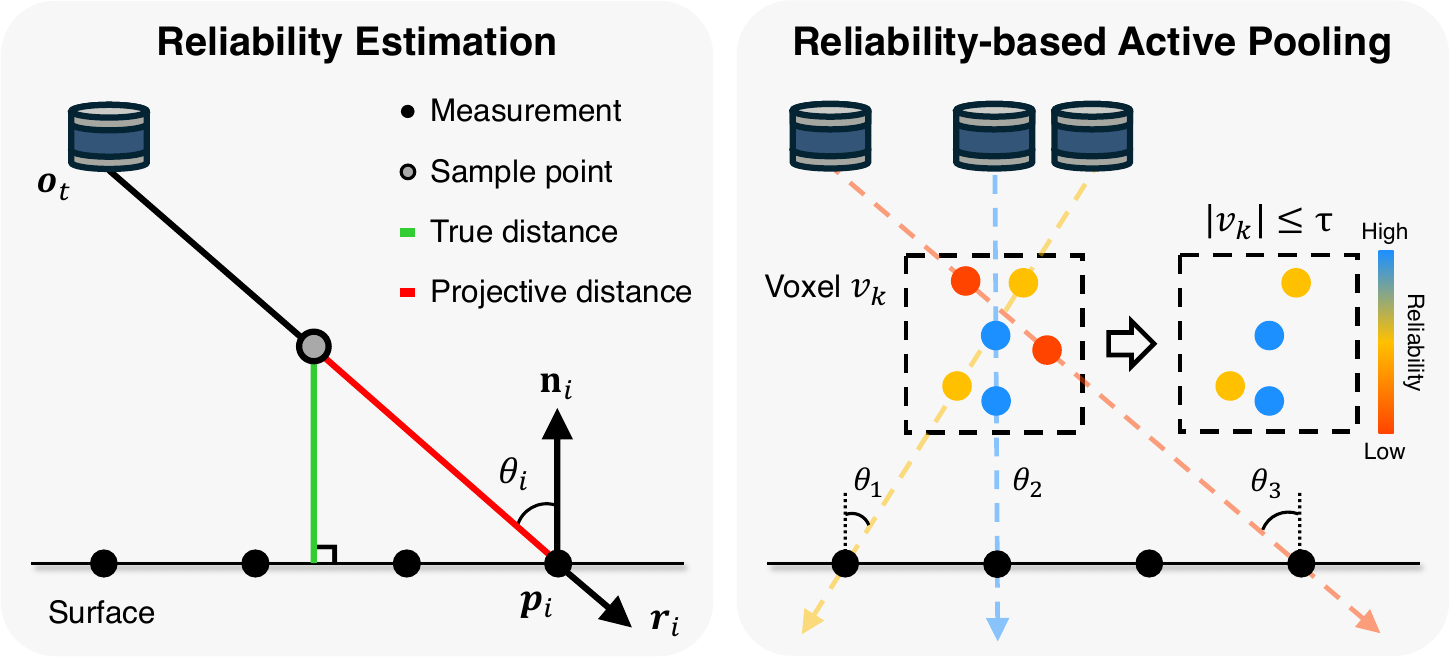}
    \caption{
    \textbf{Reliability-based active pooling.}
    Samples along each LiDAR ray are assigned to voxels, where at most $\tau$ samples are retained per voxel.
    If $\vert v_k \vert > \tau$, the $\tau$ most \textit{reliable} samples are selected by ranking the expected squared error of the projective signed distance, which increases with incidence angle and range-dependent measurement uncertainty.
    }
    \label{fig:pooling}
    % \vspace{-1.0em}
\end{figure}

Under an idealized setting with no nearby surfaces or occlusions (\cref{fig:pooling}), the projective signed distance exhibits a systematic bias that depends on the incidence angle \(\theta_i\) between the ray direction \(\vr_i\) and the surface normal \(\vn_i\), as well as the depth \(d\).
Under this assumption, we model the bias as
\begin{equation}
\label{eq:8}
   \mathrm{b} = (\| \vr_i \|_2 - d)(1-\cos\theta_i), \quad \theta_i \in [0, \pi/2).
\end{equation}
In practice, we drop the depth-dependent term to avoid systematically favoring near-range samples, yielding
\begin{equation}
\label{eq:9}
   \mathrm{b'} = 1 - \cos\theta_i.
\end{equation} 
We model measurement uncertainty as a range-dependent function following prior work~\citep{funk2021multi}:
\begin{equation}
\label{eq:10}
   \sigma_{a} = \alpha \frac{\| \vr_i \|_2}{r_p},
\end{equation}
where \(\alpha\) is a scaling parameter. 
Combining bias and variance, we define sample reliability using the expected squared error:
\begin{equation}
\label{eq:11}
   \mathrm{MSE} = \mathbb{E}[e^2] = \mathrm{b'}^2 + \sigma_{a}^2.
\end{equation}
When a voxel exceeds its capacity, we keep the \(\tau\) samples with the lowest \ac{MSE}.
This strategy limits redundancy in densely observed regions, increasing the likelihood that sparsely scanned regions contribute to training and improving surface completeness while keeping the buffer compact.

% Figure - Active Pooling
%%%%%%%%%%%%%%%%%%%%%%%%%%%%%%%%%%%%%%%%%%%%%%%%%%%%%%%%%%%
%%%%%%%%%%%%%%%%%%%%%%%%%%%%%%%%%%%%%%%%%%%%%%%%%%%%%%%%%%%

\subsection{Uncertainty-guided Active Sampling}
\label{sub:sampling}
Replay-based incremental mapping commonly uses random sampling~\citep{zhu2022nice, pan2024pin} or uniform voxel-wise sampling~\citep{song2024n}, which treats all regions equally and does not prioritize newly observed or poorly reconstructed areas.
As a result, optimization may repeatedly revisit well-converged regions, slowing learning where the geometry remains under-constrained.

We therefore propose an uncertainty-guided active sampling scheme that allocates training updates to regions where the neural distance field is most uncertain. 
Following prior work on uncertainty estimation for neural implicit representations~\citep{goli2024bayes, tao2025silvr}, we employ a perturbation field \(\mathcal{F}_\theta: \Real{3} \rightarrow \Real{3}\), parameterized by \(\theta\). 
For a query point \(\vq\), the perturbation is obtained by trilinear interpolation,
\begin{equation}
   \mathcal{F}_\theta(\vq) = \mathrm{Trilinear}(\vq, \theta). 
\end{equation}

% Intuitively, model uncertainty reflects how much geometry can be perturbed without significantly degrading reconstruction quality.
We estimate model uncertainty by approximating the covariance of the posterior \(p(\theta|\mathcal{D})\) using a Laplace approximation around the \ac{MAP} estimate \(\hat{\theta}\).
Using a second-order Taylor expansion of the loss,
\begin{equation}
\label{eq:13}
   \mathcal{L}(\mathcal{D}; \theta)
   \approx \mathcal{L}(\mathcal{D}; \hat{\theta}) + \frac{1}{2}(\theta - \hat{\theta})\tran \mathbf{H} (\theta - \hat{\theta}),
\end{equation}
the posterior is approximated as \(p(\theta|\mathcal{D}) \approx \mathcal{N}(\theta;\hat{\theta}, \mathbf{\Sigma})\), where \(\mathbf{\Sigma} = \mathbf{H}^{-1}\) and \(\mathbf{H} = \nabla^2_{\theta} \mathcal{L}(\mathcal{D; \theta}) \vert_{\hat{\theta}}\) is the Hessian at \(\hat{\theta}\)~\citep{daxberger2021laplace}.
Following~\citep{tao2024silvr}, we assume a zero-mean Gaussian prior \(p(\theta) = \mathcal{N}(\theta;0, \gamma^2\mathbf{I})\) and approximate $\mathbf{H}$ using the Fisher information matrix:
\begin{equation}
\label{eq:14}
   \mathbf{H} 
   \approx -\gamma^{-2}\mathbf{I} - \sum_{n=1}^N \mathbf{J}_n \mathbf{J}_n\tran,
\end{equation}
where \(\mathbf{J}_n\) is the Jacobian of the model output with respect to \(\theta\) and it can be efficiently computed via backpropagation.
Due to the grid structure of the perturbation field, each parameter affects only a local subset of grid cells, making the Hessian sparse. 
We therefore approximate the covariance using only diagonal entries,
\begin{equation}
\label{eq:16}
   \mathbf{\Sigma}
   \approx \mathrm{diag}\!\left(\sum_{n=1}^N \mathbf{J}_n \mathbf{J}_n^\top + \gamma^{-2}\mathbf{I}\right)^{-1}.
\end{equation}
Finally, the model uncertainty \(\sigma_e\) at a query point \(\vq\) is computed as the norm of the corresponding variance vector. %, denoted \(\sigma_e(\vq)\). %, where \(\vsigma_e=(\sigma_{x}, \sigma_y, \sigma_z)\).

% % Figure - Active Sampling
% %%%%%%%%%%%%%%%%%%%%%%%%%%%%%%%%%%%%%%%%%%%%%%%%%%%%%%%%%%%
\captionsetup[subfigure]{labelformat=empty}

\begin{figure}[t]
    \centering
    \includegraphics[width=0.99\linewidth]{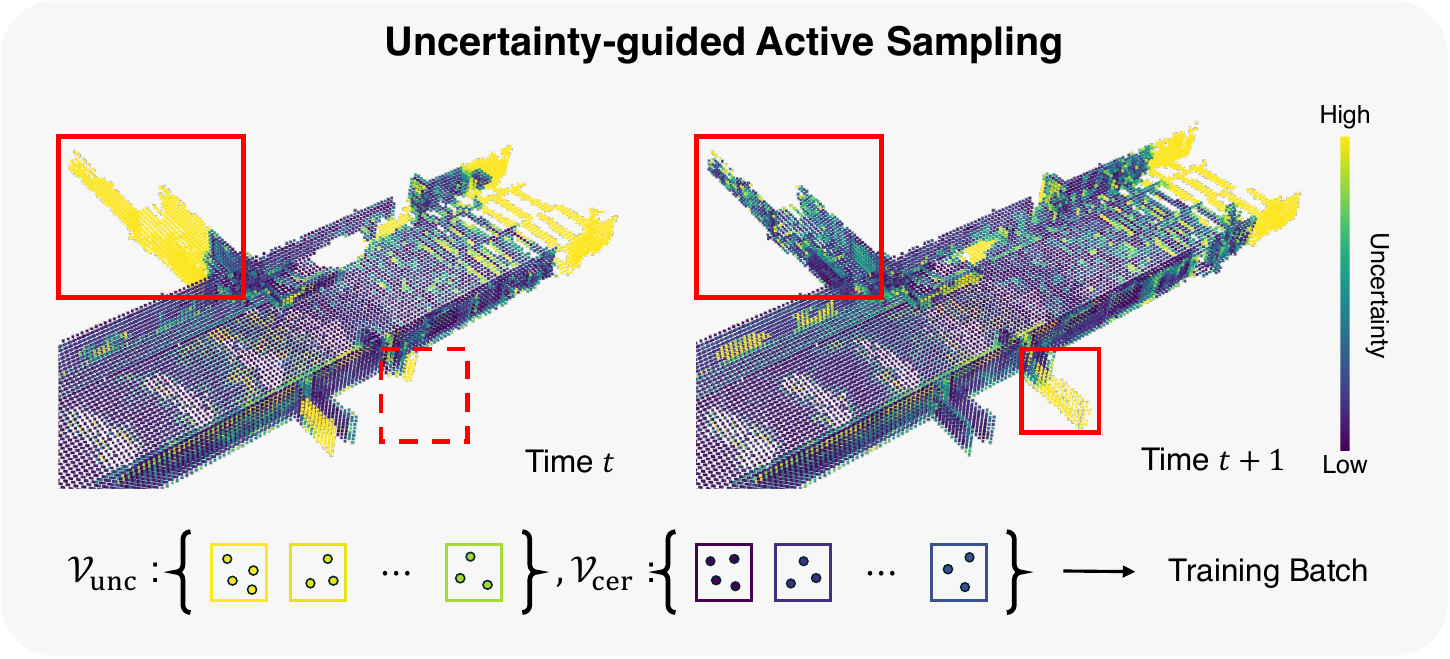}
    \caption{
    \textbf{Uncertainty-guided active sampling.} 
    We estimate model uncertainty of the neural map at each time step. 
    Red boxes illustrate how uncertainty evolves; regions become more certain as they receive supervision, while newly observed regions remain highly uncertain until sufficient updates are incorporated. 
    Guided by this uncertainty, we construct training batches by combining samples from uncertain and certain voxel sets.
    }
    % \vspace{-1.0em}
    \label{fig:sampling}
\end{figure}
% %%%%%%%%%%%%%%%%%%%%%%%%%%%%%%%%%%%%%%%%%%%%%%%%%%%%%%%%%%%

% Figure
\begin{figure*}[!t]
    \centering
    \includegraphics[width=0.99\linewidth]{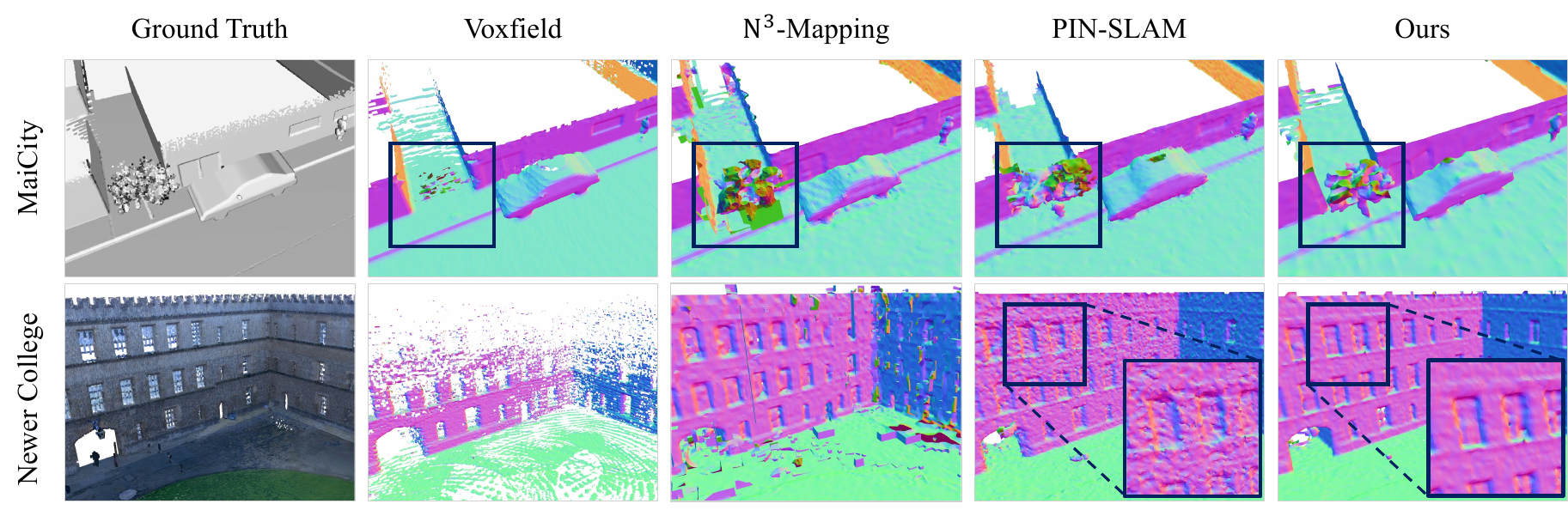}
    % \vspace{-1.5em}
    \caption{
    \textbf{Qualitative comparison on MaiCity~\cite{vizzo2021poisson} and Newer College~\cite{ramezani2020newer}.}
    Top (MaiCity): \nickname better preserves fine structures (boxed), while baselines show missing geometry, floating artifacts, and inflated shapes.
    Bottom (Newer College): \nickname produces smoother, more complete surfaces with fewer unreconstructed regions than the baselines.
    (\textit{Best viewed when zoomed in.})
    }
    \label{fig:qualitative_mai_ncd}
    % \vspace{-1.0em}
\end{figure*}

At each time step, we evaluate \(\sigma_e\) at the centers of voxels constructed from samples in the replay buffer \(\mathcal{D}_p\) (\cref{fig:sampling}).
We then divide voxels into an uncertain set $\mathcal{V}_{unc}$ and a certain set $\mathcal{V}_{cer}$ using a threshold $\lambda$:
\begin{equation}
    \mathcal{V}_{unc} = \{ v | \sigma_{e} \geq \lambda \}, \mathcal{V}_{cer} = \{ v | \sigma_{e} < \lambda \}.
\end{equation}
At each training iteration, we construct a mini-batch by drawing \(N_{unc}\) samples from \(\mathcal{V}_{unc}\) and \(N_{cer}=N-N_{unc}\) samples from \(\mathcal{V}_{cer}\).
This strategy prioritizes under-constrained regions while maintaining coverage across the rest of the map. 
Sampling exclusively from high-uncertainty voxels can over-concentrate optimization, whereas low uncertainty does not necessarily imply correct supervision, since the supervision itself may be biased. 
By balancing uncertainty and certain voxel sets, the proposed batching improves both surface coverage and training efficiency within a fixed online training budget.
\section{EXPERIMENTS}
\label{sec:exp}

\subsection{Experimental Setup}
\label{subsec:experimental_setup}
\smallskip\noindent\textbf{Implementation Details.}
All experiments were run on a single NVIDIA RTX 3080 GPU and an Intel i7-12700 CPU. 
We use a multi-resolution sparse voxel grid with $L=2$ levels and voxel sizes of \SI{0.3}{\meter} and \SI{0.45}{\meter}.
Each corner vertex stores a feature vector of dimension $D=8$, and the decoder is a lightweight \ac{MLP} with two hidden layers of 32 units.
We optimize using Adam with a learning rate of $0.01$.
To reflect online constraints, each time step performs 15 training iterations with a batch size of $N=16384$.
Along each LiDAR ray, we sample $(N_f, N_b, N_r) = (3, 1, 2)$ points with truncation distance $d_{tr}=\SI{0.3}{\meter}$.
Surface normals for incidence angle computation are estimated from the input point cloud using Open3D~\citep{zhou2018open3d}.
For active pooling, we set \(\tau=256\).
For active sampling, we use $\lambda=0.98$ and draw $N_{unc}=1000$ samples from high-uncertainty voxels and $N - N_{unc}$ samples from the rest.
Unless stated otherwise, we use the same settings across datasets.

% Table - MaiCity
\begin{table}[!t]
\setlength{\tabcolsep}{4.pt}
\centering
\caption{\textbf{Quantitative results on MaiCity~\cite{vizzo2021poisson}.} Precision, recall, and F1-score are computed at \SI{10}{\centi\meter}.
Best results are shown in \textbf{bold}, second best are \underline{underlined}.}
\label{tab:maicity}
% \begin{adjustbox}{width=1.0\linewidth}
% {
\begin{tabular}{l|ccc|ccc}
\toprule
\multicolumn{1}{l|}{Method}
& Acc. $\downarrow$ & Comp. $\downarrow$ & C-L1 $\downarrow$ & P $\uparrow$ & R $\uparrow$ & F1 $\uparrow$ \\ 
\midrule
VDBFusion~\cite{vizzo2022vdbfusion}
& \textbf{1.58} & 27.52 & 14.55 
& \textbf{99.35} & 79.48 & 88.31
\\
Voxfield~\cite{pan2022voxfield}
& \underline{2.02} & 12.12 & \underline{7.07} 
& \underline{99.05} & 86.51 & \textbf{92.35}
\\ \midrule
{\scriptsize SHINE-Mapping~\cite{zhong2023shine}}
& 5.64 & 11.93 & 8.78 
& 83.59 & 79.79 & 81.64 
\\
N$^3$-Mapping~\cite{song2024n}
& 5.63 & \underline{9.35} & 7.49  
& 88.71 & \underline{91.93} & 90.29
\\
PIN-SLAM~\cite{pan2024pin}     
& 3.49 & 11.04 & 7.27
& 93.07 & 91.62 & \underline{92.34}
\\
\nickname (Ours)                 
& 3.62 & \textbf{7.92} & \textbf{5.77}
& 92.11 & \textbf{92.46} & 92.29
\\ 
\bottomrule
\end{tabular}
% \vspace{-1em}
% }
% \end{adjustbox}
\end{table}

\smallskip\noindent\textbf{Datasets.} 
We evaluate our method on three benchmarks: MaiCity~\cite{vizzo2021poisson}, Newer College~\cite{ramezani2020newer}, and Oxford Spires~\cite{tao2025oxford}.
MaiCity is a synthetic dataset with simulated LiDAR scans, whereas Newer College and Oxford Spires are real-world datasets collected using handheld LiDAR systems.
Oxford Spires is particularly challenging due to its large scale and complex architectural structures.

\smallskip\noindent\textbf{Baselines.} 
We compare our method to three state-of-the-art neural mapping baselines: SHINE-Mapping~\cite{zhong2023shine}, N$^3$-Mapping~\cite{song2024n}, and PIN-SLAM~\cite{pan2024pin}.
In addition, we test two classical \ac{TSDF}-fusion methods: VDBFusion~\cite{vizzo2022vdbfusion} and Voxfield~\cite{pan2022voxfield}.
For fair comparison, all methods are provided with ground-truth sensor poses, and the localization module of PIN-SLAM was disabled.
We use the open-source implementations of each baseline with their recommended settings.

% Table - Newer College
\begin{table}[t!]
\setlength{\tabcolsep}{4.pt}
\centering
\caption{\textbf{Quantitative results on Newer College~\cite{ramezani2020newer}.} Precision, recall, and F1-score are computed at \SI{20}{\centi\meter}.
Best results are shown in \textbf{bold}, second best are \underline{underlined}.}
\label{tab:newer_college}
% \begin{adjustbox}{width=1.0\linewidth}
% {
\begin{tabular}{l|ccc|ccc}
\toprule
\multicolumn{1}{l|}{Method} 
& Acc. $\downarrow$ & Comp. $\downarrow$ & C-L1 $\downarrow$ & P $\uparrow$ & R $\uparrow$ & F1 $\uparrow$ \\ 
\midrule
VDBFusion~\cite{vizzo2022vdbfusion}
& \textbf{4.60} &  21.96 &  13.28
& \textbf{99.38} &  85.29 &  91.80
\\
Voxfield~\cite{pan2022voxfield}
& \underline{4.73} & 30.90 & 17.81
& 89.58 & 52.43 & 66.15
\\ \midrule
{\scriptsize SHINE-Mapping~\cite{zhong2023shine}}
& 6.94 & 11.27 & 9.10  
& \underline{95.07} & 91.95 & 93.48
\\
N$^3$-Mapping~\cite{song2024n}
& 8.49 & \underline{10.09} & 9.26  
& 89.38 & \underline{94.02} & 91.64
\\
PIN-SLAM~\cite{pan2024pin}     
& 6.87 & 10.66 & \underline{8.76}  
& 94.05 & 93.50 & \underline{93.77}
\\
\nickname (Ours)               
& 6.55 & \textbf{10.02} & \textbf{8.28}
& 94.93 & \textbf{94.04} & \textbf{94.48}
\\
\bottomrule
\end{tabular}
% \vspace{-1em}
% }
% \end{adjustbox}
\end{table}

% Figure
\begin{figure*}[!t]
    \centering
    \includegraphics[width=0.99\linewidth]{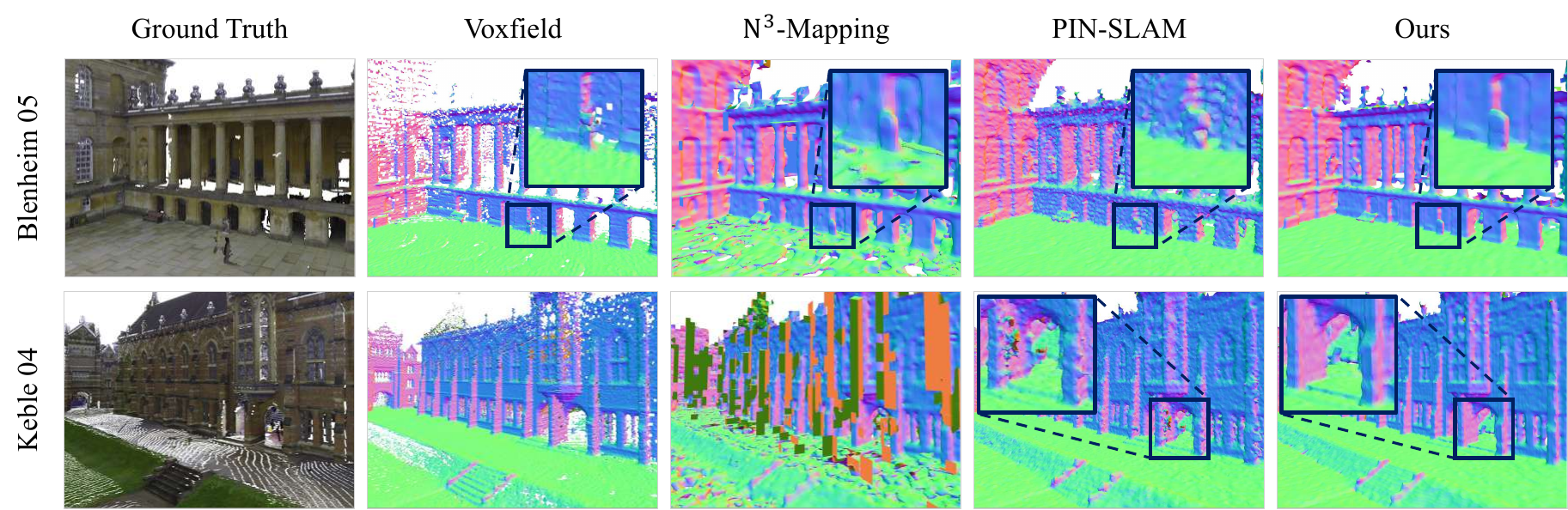}
    % \vspace{-1.5em}
    \caption{
    \textbf{Qualitative comparison on Oxford Spires~\cite{tao2025oxford}.}
    Top (\texttt{Blenheim 05}) and bottom (\texttt{Keble 04}): Baselines tend to miss geometry in under-observed regions or add spurious surfaces, such as floating artifacts, particularly around complex structures.
    In contrast, \nickname produces more spatially coherent reconstructions with improved coverage across the scene.
    (\textit{Best viewed when zoomed in.})
    }
    \label{fig:qualitative_spires}
    % \vspace{-1.0em}
\end{figure*}

% Table - Oxford Spires
\begin{table*}[!t]
\setlength{\tabcolsep}{7.5pt}
\centering
\caption{
\textbf{Quantitative results on Oxford Spires~\cite{tao2025oxford}.}
Precision, recall, and F1-score are computed at \SI{20}{\centi\meter}.
Best results are shown in \textbf{bold}, second best are \underline{underlined}.
}
\label{tab:spires}
% \begin{adjustbox}{width=0.90\linewidth}
% {
\begin{tabular}{cl|ccc|ccc}
\toprule[1.2pt]
\multicolumn{1}{c}{Sequence} &
\multicolumn{1}{l|}{Method} &
\multicolumn{1}{c}{Accuracy $\downarrow$} &
\multicolumn{1}{c}{Completeness $\downarrow$} &
\multicolumn{1}{c|}{C-L1 $\downarrow$} &
\multicolumn{1}{c}{Precision $\uparrow$} &
\multicolumn{1}{c}{Recall $\uparrow$} &
\multicolumn{1}{c}{F1 $\uparrow$} \\
\midrule[0.7pt]
\multirow{6}{*}[-0.5em]{Blenheim Palace 05}
& VDBFusion \cite{vizzo2022vdbfusion}
& \textbf{4.11} & 157.24 & 80.68
& \textbf{98.95} & 15.07 & 26.16
\\
& Voxfield \cite{pan2022voxfield}
& \underline{6.25} & 68.78 & 37.51
& \underline{94.78} & 49.14 & 64.72
\\
\cmidrule(l){2-8}
& SHINE-Mapping \cite{zhong2023shine}
& 9.72  & 68.71  & 39.22
& 88.98 & 52.38 & 65.94
\\
& N$^3$-Mapping \cite{song2024n}
& 15.25 & \textbf{46.18} & \underline{30.71}
& 68.69 & \underline{62.02} & 65.19
\\
& PIN-SLAM \cite{pan2024pin}
& 9.45  & 58.52  & 33.98
& 86.88 & 60.55 & \underline{71.36}
\\
& \nickname (Ours)
& 8.89 & \underline{49.51} & \textbf{29.20}
& 88.67 & \textbf{65.21} & \textbf{75.15}
\\
\midrule[0.7pt]
\multirow{6}{*}[-0.5em]{Christ Church 02}
& VDBFusion \cite{vizzo2022vdbfusion}
& \textbf{7.51} & 139.27 & 73.39
& \textbf{92.93} & 20.58 & 33.70
\\
& Voxfield \cite{pan2022voxfield}
& \underline{8.21} & 102.75 & 55.48
& \underline{91.44} & 38.33 & 54.02
\\
\cmidrule(l){2-8}
& SHINE-Mapping \cite{zhong2023shine}
& 10.42 & 102.99 & 56.70
& 86.33 & 37.77 & 52.55
\\
& N$^3$-Mapping \cite{song2024n}
& 14.20 & \textbf{90.54} & \underline{52.37}
& 72.19 & \underline{42.68} & 53.64
\\
& PIN-SLAM \cite{pan2024pin}
& 10.05 & 99.36  & 54.71
& 86.26 & 42.09 & \underline{56.57}
\\
& \nickname (Ours)
& 9.75 & \underline{93.15} & \textbf{51.45}
& 87.26 & \textbf{44.26} & \textbf{58.73}
\\
\midrule[0.7pt]
\multirow{6}{*}[-0.5em]{Keble College 04}
& VDBFusion \cite{vizzo2022vdbfusion}
& \textbf{6.26} & 73.54 & 39.90
& \textbf{98.12} & 53.21 & 69.00
\\
& Voxfield \cite{pan2022voxfield}
& \underline{8.01} & 33.69 & 20.85
& \underline{93.96} & 74.93 & \textbf{83.37}
\\
\cmidrule(l){2-8}
& SHINE-Mapping \cite{zhong2023shine}
& 10.78 & 34.02 & 22.40
& 86.80 & 74.68 & 80.28
\\
& N$^3$-Mapping \cite{song2024n}
& 15.62 & \textbf{26.26} & 20.94
& 67.53 & 77.40 & 72.13
\\
& PIN-SLAM \cite{pan2024pin}
& 11.72 & 29.17 & \underline{20.44}
& 81.26 & \underline{79.37} & 80.30
\\
& \nickname (Ours)
& 11.23 & \underline{26.53} & \textbf{18.88}
& 81.79 & \textbf{81.02} & \underline{82.01}
\\
\midrule[0.7pt]
\multirow{6}{*}[-0.5em]{Observatory Quarter 01}
& VDBFusion \cite{vizzo2022vdbfusion}
& \textbf{3.79} & 105.63 & 54.71
& \textbf{99.08} & 33.23 & 49.77
\\
& Voxfield \cite{pan2022voxfield}
& \underline{5.21} & 40.46 & 22.83
& \underline{97.04} & 67.52 & 79.63
\\
\cmidrule(l){2-8}
& SHINE-Mapping \cite{zhong2023shine}
& 9.27 & 40.67 & 24.97
& 88.82 & 65.88 & 75.65
\\
& N$^3$-Mapping \cite{song2024n}
& 13.86 & \textbf{20.45} & \underline{17.15}
& 72.48 & \underline{79.22} & 75.70
\\
& PIN-SLAM \cite{pan2024pin}
& 9.83  & 31.15  & 20.49
& 85.09 & 76.31 & \underline{80.46}
\\
& \nickname (Ours)
& 9.09 & \underline{22.53} & \textbf{15.81}
& 87.70 & \textbf{81.81} & \textbf{84.65}
\\
\bottomrule[1.25pt]
\end{tabular}
% \vspace{-1em}
% }
% \end{adjustbox}
\end{table*}

\smallskip\noindent\textbf{Metrics.}
Following previous work~\cite{zhong2023shine, song2024n, pan2024pin}, we evaluate mapping performance based on the quality of the surface reconstruction of the extracted meshes.
Meshes are generated by querying signed distances on a uniform 3D grid at \SI{10}{\centi\meter} resolution and extracting the surface with marching cubes~\citep{lewiner2003efficient}.
We adopt the evaluation protocol of the Oxford Spires benchmark~\cite{tao2025oxford}, reporting Accuracy (Acc.), Completeness (Comp.), and Chamfer-L1 distance (C-L1) in centimeters, along with Precision (P), Recall (R), and F1-score (F1) in percent.

\subsection{Surface Reconstruction}
\label{subsec:results}

Quantitative results are summarized in \cref{tab:maicity,tab:newer_college,tab:spires}, and qualitative comparisons are shown in \cref{fig:qualitative_mai_ncd,fig:qualitative_spires}~\footnote{Note that the Ground Truth images show rendered views from high-quality Leica TLS tripod scanners. This data was not used in the reconstructions.}.
Overall, \nickname improves surface coverage, consistently increasing recall and achieving the highest F1-scores on most datasets and sequences, with only a modest reduction in precision.

On MaiCity and Newer College, our method improves the balance between geometric fidelity and coverage, as reflected in higher F1-scores and lower Chamfer-L1 values compared to prior neural distance field-based baselines.
These improvements are also visible in \cref{fig:qualitative_mai_ncd}, where \nickname better preserves fine structures while producing smoother and more complete reconstructions.

On Oxford Spires, \nickname remains robust across all sequences, achieving high surface coverage with smooth geometry in complex scenes (\cref{fig:qualitative_spires}).
Classical \ac{TSDF}-fusion methods achieve high accuracy but low completeness, often leaving large unreconstructed regions.
Among neural baselines, N$^3$-Mapping attains the highest completeness but often introduces floating artifacts as shown in \cref{fig:qualitative_mai_ncd,fig:qualitative_spires}, which reduce accuracy and precision.
PIN-SLAM captures the overall structure well but tends to produce rough or inflated surfaces in challenging areas.
In contrast, \nickname achieves higher recall and F1-score while maintaining competitive geometric accuracy and more coherent surfaces.

\subsection{Ablation Studies}
\begin{figure}[t]
  \centering
  \vspace{-0.5em}
  \begin{subfigure}[b]{0.48\linewidth}
    \centering
    \includegraphics[width=\linewidth]{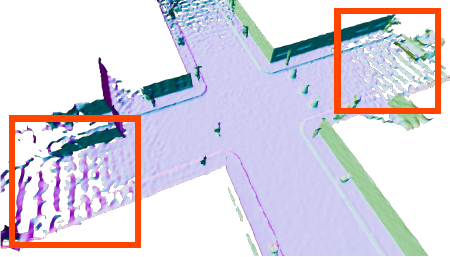}
    \caption{Without active pooling}
  \end{subfigure}\hfill
  \begin{subfigure}[b]{0.48\linewidth}
    \centering
    \includegraphics[width=\linewidth]{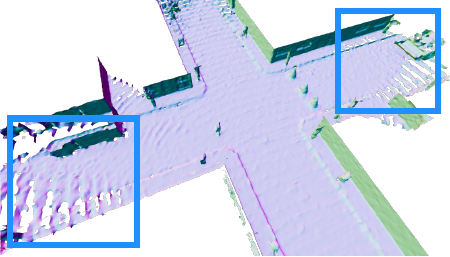}
    \caption{With active pooling}
  \end{subfigure}
  \caption{
  Effect of active pooling on MaiCity.
  Without active pooling, the reconstruction shows holes and fragmented surfaces in poorly supervised regions, whereas active pooling yields more complete geometry.
  }
  \label{fig:active_pooling_training}
  % \vspace{-0.5em}
\end{figure}

\begin{figure}[!t]
    \centering
    \includegraphics[width=0.98\linewidth]{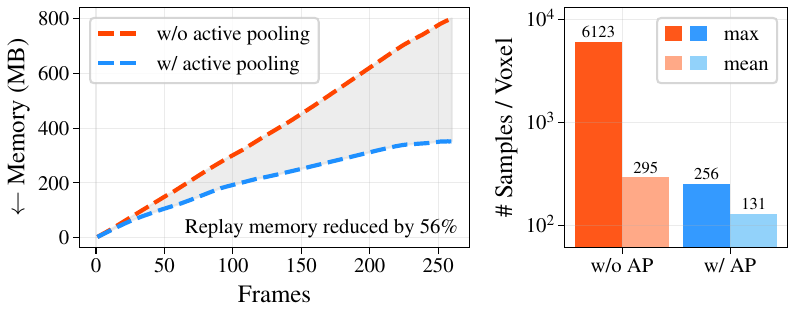}
    \vspace{-0.2em}
    \caption{
    Replay buffer memory on Newer College (260 frames).
    Without active pooling, replay memory grows approximately linearly as new samples accumulate.
    With active pooling, memory increases initially and then plateaus due to the per-voxel capacity limit.
    On this sequence, active pooling reduces replay memory by approximately 56\% and mitigates spatial sample imbalance.
    }
    \label{fig:active_pooling_memory}
    % \vspace{-0.5em}
\end{figure}

\begin{figure}[!t]
    \centering
    \includegraphics[width=0.98\linewidth]{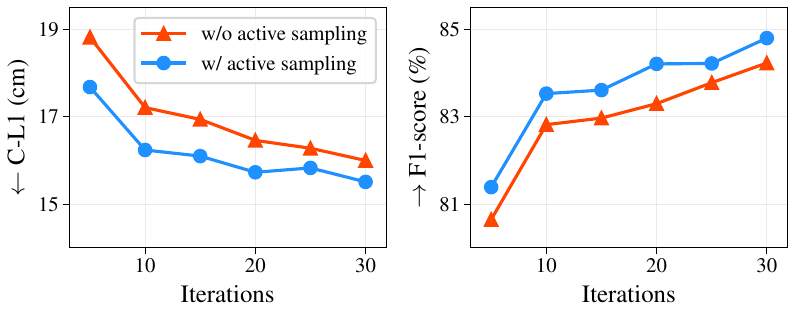}
    \vspace{-0.3em}
    \caption{
    Training efficiency of active sampling on \texttt{Observatory Quarter 01}.
    Under a fixed online training budget, active sampling achieves higher reconstruction quality in fewer optimization iterations by prioritizing samples from high-uncertainty regions.
    }
    \label{fig:active_sampling_iteration}
    % \vspace{-0.5em}
\end{figure}

In this section, we analyze the contribution of each component of \nickname through a series of ablation studies.
Quantitative results are reported in \cref{tab:ablation}, and qualitative comparisons are shown in \cref{fig:active_pooling_memory,fig:active_pooling_training,fig:active_sampling_iteration}.

% Ablation - Table
\begin{table}[t!]
\setlength{\tabcolsep}{4.pt}
\centering
\caption{
\textbf{Ablation study on Observatory Quarter 01.}
Best results are shown in \textbf{bold}, second best are \underline{underlined}.
}
\label{tab:ablation}
% \begin{adjustbox}{width=1.0\linewidth}
% {
\begin{tabular}{c|c|ccc|ccc}
\toprule
A.P. & A.S.
& Acc. $\downarrow$ & Comp. $\downarrow$ & C-L1 $\downarrow$ & P $\uparrow$ & R $\uparrow$ & F1 $\uparrow$ \\ 
\midrule
\(\times\) & \(\times\)         
& \textbf{8.85} & 25.03 & 16.94 & \underline{87.59} & 78.80 & 82.96 \\
\checkmark (R) & \(\times\)  
& 9.18 & \underline{22.77} & \underline{15.98} & 87.28 & 81.61 & 84.35 \\
\checkmark ($\sim$A) & \(\times\)
& 9.13 & 22.92 & 16.02 & 87.39 & 81.21 & 84.19 \\
\checkmark (A) & \(\times\)
& 9.18 & 22.79 & \underline{15.98} & 87.32 & \underline{81.74} & \underline{84.44} \\
\(\times\) & \checkmark        
& \underline{8.98} & 23.23 & 16.10 & 87.38 & 80.17 & 83.62 \\
\checkmark (A) & \checkmark 
& 9.09 & \textbf{22.53} & \textbf{15.81} & \textbf{87.70} & \textbf{81.81} & \textbf{84.65} \\ 
\bottomrule
\end{tabular}
% }
% \end{adjustbox}
\begin{flushleft}
\footnotesize
A.P. and A.S. denote active pooling and active sampling. For active pooling variants, R denotes random selection, and $\sim$A and A denote selection of the most unreliable and reliable samples, respectively.
\end{flushleft}
% \vspace{-1.5mm}
\end{table}

% Ablation - Active Pooling
\captionsetup[subfigure]{labelformat=parens}

\captionsetup[subfigure]{labelformat=empty}

% Ablation - Active Pooling Memory
% memory efficiency

\smallskip\noindent\textbf{Effect of Active Pooling.} 
Reliability-based active pooling significantly improves reconstruction completeness.
As shown in \cref{tab:ablation}, it improves recall over the variant without pooling.
\Cref{fig:active_pooling_training} further shows that active pooling reduces holes and fragmented surfaces in less frequently observed regions, yielding smoother and more complete reconstructions.
This improvement is consistent with reduced spatial imbalance in the replay buffer, as active pooling prevents densely observed regions from dominating replay (\cref{fig:active_pooling_memory}).

\smallskip\noindent\textbf{Comparison between Pooling Strategies.} 
We further compare the proposed reliability-based pooling with random selection under the same voxel-wise capacity constraint.
As shown in \cref{tab:ablation}, active pooling yields modest but consistent gains over random selection and selection of the most unreliable samples, indicating that the estimated supervision error provides a useful signal for replay retention.

\smallskip\noindent\textbf{Memory Efficiency of Active Pooling.} 
The effect of active pooling on memory scalability is illustrated in \cref{fig:active_pooling_memory}. 
Without active pooling, replay memory grows monotonically as all incoming samples are retained.
By bounding the number of stored samples per voxel, active pooling slows the growth of replay memory and reduces it by more than 50\% on Newer College, while also mitigating spatial sample imbalance.

% Ablation - Active Sampling

\smallskip\noindent\textbf{Effect of Active Sampling.} 
We next evaluate the contribution of uncertainty-guided active sampling.
As shown in \cref{tab:ablation}, active sampling improves recall and F1-score by focusing updates on poorly constrained regions rather than repeatedly revisiting well-converged areas.
\Cref{fig:active_sampling_iteration} further shows that active sampling reaches higher reconstruction quality in fewer optimization iterations, indicating improved training efficiency under a limited optimization budget.

\smallskip\noindent\textbf{Combined Effect of Both Modules.}
Finally, combining active pooling and active sampling yields the best overall reconstruction quality. 
The full pipeline achieves the lowest Chamfer-L1 and the highest F1-score in \cref{tab:ablation}, showing that effective pooling and sampling policies improve incremental neural mapping.

\begin{table}[!t]
\centering
\caption{
\textbf{Runtime comparison on MaiCity~\citep{vizzo2021poisson} and Newer College~\citep{ramezani2020newer}.}
We report the average processing time per frame in seconds and frame rate in hertz for each method.
}
\label{tab:runtime_comparison}
\begin{tabular}{lcccc}
\toprule
Dataset $\rightarrow$ &
  \multicolumn{2}{c}{MaiCity~\citep{vizzo2021poisson}} &
  \multicolumn{2}{c}{Newer College~\citep{ramezani2020newer}} \\ 
  \cmidrule(lr){2-3} \cmidrule(lr){4-5}
Method $\downarrow$ &
  \multicolumn{1}{c}{Time $\downarrow$} &
  \multicolumn{1}{c}{FPS $\uparrow$} &
  \multicolumn{1}{c}{Time $\downarrow$} &
  \multicolumn{1}{c}{FPS $\uparrow$} \\ 
\midrule
{\scriptsize SHINE-Mapping~\cite{zhong2023shine}} &
1.48 & 0.68 & 2.65 & 0.38 \\
N$^3$-Mapping~\cite{song2024n} &
1.35 & 0.74 & 5.77 & 0.17 \\
PIN-SLAM~\cite{pan2024pin} &
0.11 & 9.09 & 0.10 & 9.80 \\
\nickname (Ours) &
0.13 & 7.69 & 0.10 & 9.80\\
\bottomrule
\end{tabular}
\end{table}
\begin{table}[!t]
\setlength{\tabcolsep}{5.pt}
\centering
\caption{
\textbf{Module-wise runtime analysis on Oxford Spires~\citep{tao2025oxford}.}
We report the average per-frame runtime in milliseconds of each component in our pipeline.
}
\label{tab:runtime_modules}
\begin{tabular}{lccccc}
\toprule
Sequence & Preproc. & A.P. & A.S. & Optim. & Total \\
\midrule
Blenheim Palace 05 & 32.38 & 5.81 & 10.76 & 38.19 & 87.13 \\
Christ Church 02 & 41.26 & 9.27 & 13.96 & 37.66 & 102.15 \\
Keble College 04 & 39.38 & 11.99 & 18.55 & 38.64 & 108.56 \\
Observatory Quarter 01 & 39.22 & 10.96 & 18.01 & 40.13 & 108.32 \\
\bottomrule
\end{tabular}
\begin{flushleft}\footnotesize
Preproc., A.P., A.S., and Optim. denote preprocessing, active pooling, active sampling, and map optimization, respectively.
\end{flushleft}
% \vspace{-0.5em}
\end{table}

\subsection{Runtime Analysis}
We report runtime in \cref{tab:runtime_comparison,tab:runtime_modules}.
\Cref{tab:runtime_comparison} compares the average per-frame processing time of \nickname with online neural mapping baselines on MaiCity and Newer College, and \cref{tab:runtime_modules} provides a module-wise breakdown on Oxford Spires.
\nickname achieves practical runtime for online mapping, remaining comparable to PIN-SLAM and substantially faster than SHINE-Mapping and N$^3$-Mapping.
The module-wise analysis further shows that the added overhead of active pooling and sampling remains moderate.
\section{CONCLUSION}

This paper presented \nickname, a replay management framework for incremental LiDAR mapping based on neural distance fields.
\nickname improves replay efficiency and reconstruction fidelity through two complementary components: reliability-based active pooling, which stores the most reliable samples under a per-voxel limit to reduce memory usage and sample imbalance in the replay buffer, and uncertainty-guided active sampling, which allocates the online optimization budget toward under-constrained regions for more efficient and effective training.
Across synthetic and real-world benchmarks, \nickname yields more complete reconstructions while maintaining competitive geometric accuracy.
For example, on Oxford Spires (Blenheim Palace 05), it improves recall by 4.66 pp and F1-score by 3.79 pp over PIN-SLAM.

%%% REFERENCES
{%\balance
\small
\bibliographystyle{IEEEtranN}
\bibliography{RPM_packages/RPM_string,references}
}

\end{document}